\pdfoutput=1
 
\documentclass{article}

\usepackage[numbers, sort&compress]{natbib}

\usepackage[preprint]{neurips_2025}

\usepackage[utf8]{inputenc} 
\usepackage[T1]{fontenc}    
\usepackage{graphicx}
\usepackage{url}            
\usepackage{booktabs}       
\usepackage{amsfonts}       
\usepackage{nicefrac}       
\usepackage{microtype}      
\usepackage{amsmath}
\usepackage{enumitem}
\usepackage[dvipsnames,table,xcdraw]{xcolor}
\usepackage{todonotes}
\usepackage{pifont} 
\usepackage{multirow}
\usepackage{amssymb}
\usepackage{caption}
\usepackage{subcaption}
\usepackage{wrapfig}  
\usepackage{lipsum}   
\usepackage{hyperref}       

\title{Graph-MLLM: Harnessing Multimodal Large Language Models for Multimodal Graph Learning}

\author{%
\textbf{Jiajin Liu\textsuperscript{1}, Dongzhe Fan\textsuperscript{1}, Jiacheng Shen\textsuperscript{1}, Chuanhao Ji\textsuperscript{2}} \\
\textbf{Daochen Zha\textsuperscript{3}, Qiaoyu Tan\textsuperscript{1}} \\
\textsuperscript{1}New York University Shanghai \\
\textsuperscript{2}University of Chinese Academy of Sciences \quad \textsuperscript{3}Rice University \quad \\
\texttt{\{jiajinliu, df2362, js12556, qiaoyu.tan\}@nyu.edu} \quad \\
\texttt{jichuanhao22@mails.ucas.ac.cn} \quad 
\texttt{daochen.zha@rice.edu}
}

\begin{document}

\maketitle

\begin{abstract}
Multimodal Large Language Models (MLLMs) have demonstrated remarkable capabilities in representing and understanding diverse modalities. However, they typically focus on modality alignment in a pairwise manner while overlooking structural relationships across data points. Integrating multimodality with structured graph information (i.e., multimodal graphs, MMGs) is essential for real-world applications such as social networks, healthcare, and recommendation systems. Existing MMG learning methods fall into three paradigms based on how they leverage MLLMs: Encoder, Aligner, and Predictor. MLLM-as-Encoder focuses on enhancing graph neural networks (GNNs) via multimodal feature fusion; MLLM-as-Aligner aligns multimodal attributes in language or hidden space to enable LLM-based graph reasoning; MLLM-as-Predictor treats MLLMs as standalone reasoners with in-context learning or fine-tuning. Despite their advances, the MMG field lacks a unified benchmark to fairly evaluate across these approaches, making it unclear what progress has been made. To bridge this gap, we present \textit{Graph-MLLM}, a comprehensive benchmark for multimodal graph learning by systematically evaluating these three paradigms across six datasets with different domains. Through extensive experiments, we observe that jointly considering the visual and textual attributes of the nodes benefits graph learning, even when using pre-trained text-to-image alignment models (e.g., CLIP) as encoders. We also find that converting visual attributes into textual descriptions further improves performance compared to directly using visual inputs. Moreover, we observe that fine-tuning MLLMs on specific MMGs can achieve state-of-the-art results in most scenarios, even without explicit graph structure information. We hope that our open-sourced library will facilitate rapid, equitable evaluation and inspire further innovative research in this field. The benchmark code is available at \url{https://github.com/oamyjin/Graph-MLLM}.
\end{abstract}

%

\section{Introduction}
\label{sec:intro}

\begin{figure}[htbp]
\centering
\includegraphics[width=1\textwidth]{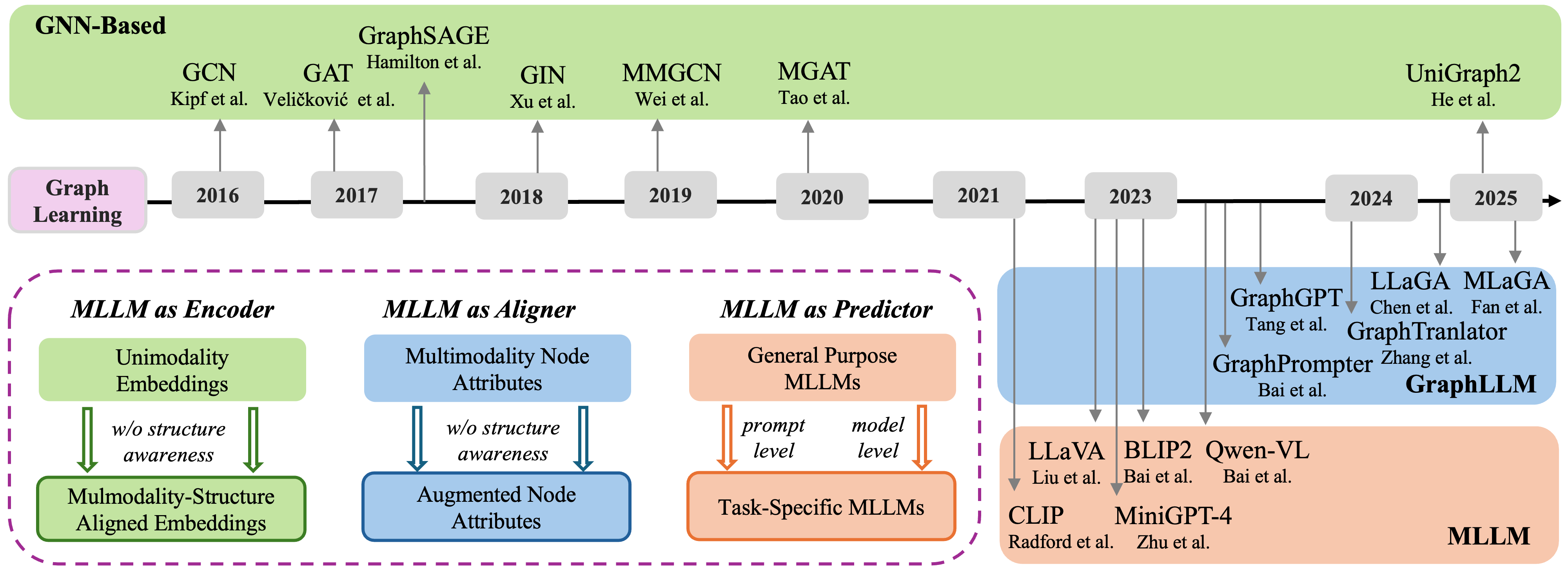}
\caption{The overview of the Graph-MLLM benchmark with the timeline of graph learning research. Existing graph learning methods are categorized into three groups based on the prediction backbone. The bottom left corner illustrates the role MLLMs play in each category in the benchmark.}
\label{fig:graphmllm}
\end{figure}

In recent years, Multimodal Large Language Models (MLLMs) have made impressive strides in learning and interpreting information across multiple modalities. Some models~\cite{clip, blip, li2023blip, girdhar2023imagebind} are pre-trained for multimodal alignment, such as CLIP~\cite{clip}, and excel at fusing image and text information for representation learning. Some other models~\cite{bai2023qwenvlversatilevisionlanguagemodel, liu2023llava, openai2024gpt4technicalreport, geminiteam2025geminifamilyhighlycapable}, such as Qwen-VL~\cite{bai2023qwenvlversatilevisionlanguagemodel}, are designed to process multimodal inputs, combining both text and visual information to enhance understanding and reasoning capabilities. However, these models generally align modalities in pairs, neglecting the underlying graph structural relationships between data points.

Multimodal graph learning (MMGL), which integrates these heterogeneous modalities with structural information, holds transformative potential by producing richer representations that enable more accurate predictions, improved clustering, and better recommendations \cite{zhu2024multimodalgraphbenchmark, yan2024graph}. Existing MMGL methods can be classified into three categories based on how MLLMs are leveraged in MMGL: Encoder, Aligner, and Predictor, as shown in Figure~\ref{fig:graphmllm}. (1) For GNN-based methods~\cite{GNN, kipf2016gcn, hamilton2017graphSAGE, velivckovic2017gat, Chien2022, uniGraph2}, MLLMs encode visual and textual attributes using pre-trained multimodal alignment models to generate node representations, which are then processed by GNNs to integrate structural information for downstream tasks; (2) For LLM-based methods~\cite{chen2024llaga, liu2024graphPrompter, tang2024graphgpt, he2024unigraph, fang2024gaugllm, sun2025graphiclunlockinggraphlearning, mlaga2025}, MLLMs can be used to align multimodal inputs into unified textual representations~\cite{blip, li2023blip, minigpt4, guo2023imagestextualpromptszeroshot} in language or latent space, enabling LLMs to perform graph tasks such as node classification and link prediction. (3) For MLLM-based methods~\cite{liu2023llava, bai2023qwenvlversatilevisionlanguagemodel, openai2024gpt4technicalreport, geminiteam2025geminifamilyhighlycapable}, MLLM is used directly as a predictor for graph tasks, harnessing their underlying multimodal understanding capabilities on MMG. 

However, the field lacks a unified benchmark to fairly evaluate across these approaches, making it unclear what progress has been made.
Some efforts~\cite{zhu2024multimodalgraphbenchmark, yan2024graph} have evaluated the potential of MLLM-as-Encoder for GNN-based methods in MMGs, but they remain confined to traditional GNN approaches and overlook emerging LLM-based graph methods that represent the current state-of-the-art~\cite{jin2024largelanguagemodelsgraphs, surveyllmforgraph,zhang2025trustglm,li2024glbench}. Moreover, they fail to provide a comprehensive evaluation of MLLMs used directly as graph predictors through either structured or unstructured supervised fine-tuning. Without fair comparisons across a diverse set of graph models, including MLLM-as-Encoder for GNN-bases methods, MLLM-as-Aligner for LLM-based methods, and MLLM-as-Predictor approaches, it remains challenging to assess the limitations of existing approaches and to identify promising avenues for multimodal graph learning research.

To address the gap, we propose \textbf{Graph-MLLM}, a comprehensive benchmark for multimodal graph learning. First, we investigate GNN-based methods~\cite{kipf2016gcn, hamilton2017graphSAGE, mmgcn, tao2020mgat} by incorporating MLLM pre-trained alignment models, such as CLIP~\cite{clip}, and state-of-the-art multimodal graph foundation models, such as UniGraph2~\cite{uniGraph2}, to generate embeddings for GNNs. Second, we evaluate LLM-based methods in MMGs by utilizing MLLMs to transform multimodal inputs into unified textual representations, and also the state-of-the-art method MLaGA~\cite{mlaga2025}. Finally, we assess the performance of general MLLMs, including Qwen-VL~\cite{bai2023qwenvlversatilevisionlanguagemodel} and LLaVA~\cite{liu2023llava}, as standalone predictors, utilizing both prompt-based and fine-tuning strategies to perform graph tasks with multimodal inputs. Our key contributions are summarized below:


\begin{itemize}[noitemsep,leftmargin=*,label=$\star$]
\item \textbf{Comprehensive Benchmark}. Graph-MLLM provides a unified experimental framework for a fair comparison of state-of-the-art graph learning methods, including GNN-based, LLM-based, and MLLM-based approaches, in five popular multimodal graph datasets. Notably, our benchmark highlights the most promising solution and offers guidance for future directions in multimodal graph research.
\item \textbf{Multi-Dimensional Analysis}. We systematically evaluate existing graph methods from multiple perspectives, including domain-specific and structure-aware multimodal alignment, the integration of visual attributes via data augmentation, and task-specific supervised fine-tuning with or without structural knowledge. \textit{\textbf{Our Key Findings:}} \textit{i)} Without MLLM enhancement, GraphLLMs outperform both GNN-based and MLLM-based methods in MMGL tasks. \textit{ii)} MLLM-as-Predictor approaches deliver the most significant performance gains, demonstrating their strong potential as the backbone model for MMGL. \textit{iii)} The sparse graph structures and imbalanced multimodal quality heavily influence the structure information effectiveness in MMGL.

\item \textbf{Open-sourced benchmark library}. We have made our benchmark library publicly available on Github, aiming to facilitate future research endeavors. We have also outlined potential future directions based on our benchmark findings to inspire further investigations.

\end{itemize}


\section{Formulations and Background}
\label{sec:preliminary}
In this section, we first formally define multimodal graphs and then discuss existing methods for multimodal graph learning tasks.

\subsection{Multimodal Graph Definitions}
Formally, a multimodal graph (MMG) can be represented as $\mathcal{G} = (\mathcal{V}, \mathcal{A}, \mathcal{T}, \mathcal{I}, \mathcal{Y})$, where $\mathcal{V}$ is the set of nodes and $\mathcal{A}$ is the adjacency matrix. $\mathcal{T}$ and $\mathcal{I}$ represent the sets of textual descriptions and images associated with nodes $\mathcal{V}$, respectively. Each node $v_i \in \mathcal{V}$ is associated with a textual attribute $\mathcal{T}_{v_i} \in \mathcal{T}$ and an image attribute $\mathcal{I}_{v_i} \in \mathcal{I}$. $\mathcal{Y}$ denotes the label space, where a node $v_i \in \mathcal{V}$ may have a label $\mathcal{Y}_{v_i} \in \mathcal{Y}$ for classification tasks. 
In this paper, we focus on node classification tasks for MMGs. Formally, given a multimodal graph $\mathcal{G}$, where the label $\mathcal{Y}_{v_i}$ of a node $v_i \in \mathcal{V}$ is missing, the task goal is to accurately predict its label.

\subsection{Graph Learning Methods}
\label{sec:graphmethods}

\vspace{3pt}
\noindent\textbf{GNN-based Methods.} 
GNNs are widely used in graph learning tasks due to their ability to aggregate information from neighboring nodes to learn expressive node embeddings. The update rule for a node embedding $\mathbf{h}_v$ in each GNN layer can be formulated as:
\begin{equation}
\mathbf{h}v^{(l+1)} = \psi\left(\phi\left({\mathbf{h}{v'}^{(l)} : v' \in \mathcal{N}(v)}\right), \mathbf{h}_v^{(l)}\right),
\end{equation}
where $\mathcal{N}(v)$ denotes the set of neighbors of node $v$, and $\phi(\cdot)$ and $\psi(\cdot)$ are aggregation and update functions, respectively. By stacking $L$ GNN layers, the final node embeddings can be utilized for downstream tasks such as node classification and link prediction.


\vspace{3pt}
\noindent\textbf{LLM-based Methods (GraphLLMs).}
LLMs play a critical role in graph learning by leveraging their natural language understanding and generation capabilities to handle graph-based tasks. To integrate graph data into the LLM embedding space, some GraphLLMs \cite{tang2024graphgpt, he2024unigraph} adopt a projection-based approach, where a projector maps node embeddings into the LLM’s token space. The LLM then processes these projected embeddings as input tokens. This process can be represented as:
\begin{equation}
\mathbf{z}_v = \text{Projector}(\mathbf{h}_v) \in \mathbb{R}^{d_{LLM}},
\end{equation}where $\mathbf{h}_v$ is the GNN-generated node embedding, and $\mathbf{z}_v$ is the embedding projected into the LLM token space.
Alternatively, some other GraphLLMs \cite{chen2024llaga, ye2023language} adopt instruction-based prompting methods that convert graph data into text sequences, enabling LLMs to infer graph-related insights through instruction-based learning. 


\vspace{3pt}
\noindent\textbf{Multimodal Large Language Models (MLLMs).}
Recent advancements in MLLMs have created new opportunities for graph learning tasks, leveraging their inherent ability to align and integrate multimodal data. Typically, a general-purpose MLLM predicts the response answer $\mathcal{R}$ based on the instruction prompt $\mathcal{P}$ and the multimodal input $\mathcal{M}$, formulated as:
\begin{equation}
\mathcal{R} = MLLM(\mathcal{P}, \mathcal{M}; \theta)
\end{equation}
where $\theta$ represents the model parameters. Similar to LLMs, the training objective minimizes the negative log-likelihood of predicting the next token in the response:
\begin{equation}
\mathcal{L}(\theta) = -\sum_{i=1}^N \log p(\mathcal{R^T}_i|\mathcal{P}, \mathcal{R^T}_{<i}; \theta)
\end{equation} where $\mathcal{R^T}$ the ground truth response and $N$ is the length of $\mathcal{R^T}$.


\section{Benchmark Design}
\label{sec:eval_frame}
In this section, we present the experimental design for the three categories of existing methods. Figure \ref{fig:graphmllm} shows our overall benchmark design, along with the timeline of the graph learning research.

\subsection{MLLM-as-Encoder for GNN-based Methods} 
\label{sec:enhancer}
As mentioned in Sec~\ref{sec:graphmethods}, GNN-based methods use encoded node features as input and then employ a GNN to integrate graph structure into the prediction process. Our goal is to evaluate both the effectiveness of the modality feature encoders and the performance of the GNN-based prediction models.

\textbf{Feature Encoders.}
To evaluate the impact of modality alignment and structural awareness at the embedding level, we adopt pre-trained multimodal models (PMMs) as node encoders. We consider three variants for comparison:

$\bullet$
The original PMM is used without any task-specific tuning, with CLIP adopted as the representative model by concatenating its text and image embeddings. To evaluate the impact of different modality settings, we also compare text-only and image-only embeddings.

$\bullet$ 
A PMM with modality alignment is then fine-tuned on specific MMGs, referred to as PMM-F, to better adapt to the target graph domain. We follow the same training paradigm as CLIP~\cite{clip}, which adopts a contrastive training objective to maximize the cosine similarity between the image and text embeddings.

$\bullet$ 
A structure-aware encoder (PMM-F-S) is further fine-tuned within a GNN framework to capture the relationships among data points. For each central node $v_i$, we randomly sample $m$ nodes from its 1-hop neighborhood. To effectively align various modalities with the structure signal, we design a structure-aware contrastive training objective:
\begin{equation}
    \mathcal{L}_v = -\log \frac{\exp \left( \text{sim}(\mathcal{E}_{TI}^{v_i}, \mathcal{E}_{TI}^{v_j}) / \tau \right)}{\sum\limits_{v_k \in \mathcal{B}} \exp \left( \text{sim}(\mathcal{E}_{TI}^{v_i}, \mathcal{E}_{TI}^{v_k}) / \tau \right)}
\end{equation}
where $\mathcal{E}_{TI}^{v_i}$ denotes the concatenated text and image embedding of the central node, $\mathcal{E}_{TI}^{v_j}$ is the multimodal embedding of the neighbors, and $\tau$ indicates the temperature parameter.

\textbf{GNN Models.} 
To assess how various GNN architectures perform with the aforementioned encoders, we evaluate representative GNN-based prediction models, including GCN~\cite{kipf2016gcn}, GraphSAGE\cite{hamilton2017graphSAGE}, MMGCN~\cite{mmgcn}, and MGAT~\cite{tao2020mgat}. Moreover, we include UniGraph2~\cite{uniGraph2}, a state-of-the-art multimodal graph foundation model that simultaneously captures modality-specific features and graph structure. As a basic baseline, we also apply a simple MLP~\cite{rosenblatt1958perceptron} model that does not incorporate structural information.

\subsection{MLLM-as-Aligner for LLM-based Methods}
When applying LLM-based methods (GraphLLMs) to MMGL, a key challenge is their inherent limitation in handling multimodal inputs, particularly visual data, as they are primarily designed for text-based tasks and rely on textual modalities. We draw on the paradigm of augmenting visual information into textual data to align multimodalities, as demonstrated in MLLM works~\cite{blip, li2023blip, minigpt4, guo2023imagestextualpromptszeroshot}. This experiment aims to evaluate the effectiveness of alignment methods and the downstream performance of GraphLLMs.

\textbf{Modality Aligner.} 
Most GraphLLMs incorporate only the textual attributes of nodes into their prompts, but with a modality aligner, visual information can also be seamlessly integrated into the input. Furthermore, by considering the structural information of the graph, visual features from neighboring nodes can be included to enhance the input through structure-aware augmentation. To evaluate the effectiveness of the modality aligner, we consider two variants: one with structure awareness and one without.

$\bullet$
At the prompt level, we leverage MLLMs to transform visual attribute $I$ of each node into a generated image description, denoted as $\mathcal{T}^I$. Given the GraphLLM's context length limitations, we summarize the original textual description \( \mathcal{T} \) and the generated image description \( \mathcal{T}^I \) into a concise summary \( \mathcal{T}^S \). The summary is used in the prompt, and its embedding $\mathcal{E}_{T^S}$ is also provided as input. The modality augmentation process is defined as:
\begin{equation}
\label{equation:generate}
\mathcal{T^I} = MLLM(\mathcal{P_\text{Generation}}, \mathcal{I}; \theta)
\end{equation} 
\begin{equation}
\label{equation:summary}
\mathcal{T}^S = MLLM(\mathcal{P}_\text{Summary}, (\mathcal{T} + \mathcal{T}^I); \theta)
\end{equation} 
where $\mathcal{P}_\text{Generation}$ and $\mathcal{P}_\text{Summary}$ are task-specific instruction prompts, and $\theta$ denotes the MLLM model parameters. 


$\bullet$
To integrate multimodal and structural information, we augment not only the anchor node’s visual input but also that of its neighbors. The summarization then yields a structure-aware, multimodal textual summary $\mathcal{T}^{SS}$ for the prompt and its embedding $\mathcal{E}_{T^{SS}}$ as feature input. This process is defined as:
\begin{equation}
\label{equation:summary_neighbor}
\mathcal{T}^{SS} = MLLM(\mathcal{P}_\text{Summary}, (\mathcal{T} + \mathcal{T}^I + \sum_{v' \in \mathcal{N}(v)} \mathcal{T}^I_{v'}); \theta)
\end{equation}
where $\mathcal{N}(v)$ denotes the set of neighbor nodes of the anchor node $v$, and $\mathcal{T}^I_{v'}$ represents the generated image description for neighbor node $v'$. 

All prompts for augmentation and graph learning tasks are detailed in Appendix.\ref{appendix:prompt}.

\textbf{GraphLLM Models.} 
To assess how various GraphLLM models perform with the aforementioned augmentation approaches, we evaluate representative GraphLLMs considering distinct architecture paradigms, including GraphPrompter~\cite{liu2024graphPrompter}, LLaGA~\cite{chen2024llaga}, GraphGPT~\cite{tang2024graphgpt}, GraphTranslator~\cite{zhang2024graphtranslator}, and state-of-the-art MLaGA~\cite{mlaga2025}. We employ CLIP~\cite{clip} as the encoder and the state-of-the-art MLLM model QWen-VL~\cite{bai2023qwenvlversatilevisionlanguagemodel} serving as the augmented.

\subsection{MLLM-as-Predictor for MLLM-based Methods}
MLLMs offer advanced modality understanding and reasoning capabilities, presenting new opportunities for enhancing performance in MMGL tasks. We include a state-of-the-art model with an MLLM backbone designed for MMGL. Additionally, to evaluate the potential of exiting MLLM for graph-structure data, we conduct assessments at both the prompt level and the model fine-tuning level.

\textbf{Prompt Level.} 
We perform zero-shot evaluation with both non-structure-aware and structure-aware approaches to benchmark MLLMs in MMGL tasks. In this setup, the MLLM predicts the label directly without task-specific training.

$\bullet$
For the non-structure-aware baseline, we construct a prompt that includes the textual description and corresponding visual input for each test node.

$\bullet$
To evaluate the MLLM’s structure-awareness, we construct a prompt that includes the top-k hop-1 neighbors for each anchor node, incorporating both the textual and visual information of the neighbors.

\textbf{Model Fine-tuning.} 
We apply instruction tuning to the MLLM using both non-structure-aware and structure-aware strategies. To evaluate the effectiveness of incorporating neighbor structure information, we employ two strategies to enhance the model's structural context during structure-aware fine-tuning. 

$\bullet$
The model fine-tuned without structure awareness performs instruction tuning at the individual node level, utilizing each node's multimodal input and ground truth for fine-tuning. 

$\bullet$
When fine-tuning with structure awareness, we enhance the prompt with structural information, enriching the anchor node's multimodal data by incorporating two strategies: (1) text-only information, and (2) text-image information from its top-k neighbors.

\textbf{MLLM Models.} 
For the MLLM-as-predictor approach in MMGL tasks, we include two representative MLLMs: Qwen-VL-7B model ~\cite{bai2023qwenvlversatilevisionlanguagemodel} and LLaVA-1.5-7B model ~\cite{liu2023llava}. We adhere to the official guidelines for LORA fine-tuning strictly.

\section{Experiments}
\label{sec:exp_analysis}
In this section, we conduct extensive experiments to answer the following research questions (\textbf{RQs}): \\
\textbf{\textit{RQ1:}} What is the performance of various GNN-based methods with pre-trained multimodal models as encoders in multimodal graphs? 
\textbf{\textit{RQ2:}} How well do state-of-the-art MLLM-based models perform, and how effective are MLLMs as aligners in transforming multimodal data into a unimodal form, enabling GraphLLMs to handle multimodal graphs? 
\textbf{\textit{RQ3:}} How effectively can general-purpose MLLMs be applied as standalone predictors to multimodal graph learning tasks? 
\textbf{\textit{RQ4:}} Which role of MLLMs holds the greatest potential for improving performance in multimodal graph learning?


\subsection{Datasets}
The datasets utilized in this benchmark cover six different datasets from the Amazon co-purchase networks~\cite{jure2014snap} and Reddit platform~\cite{hamilton2017graphSAGE}, as shown in Table~\ref{table:dataset}. They vary in scale and density, with a wide range of the number of nodes and different average degrees. Each product is represented as a node and is associated with multimodal information: textual description and visual image from the product or post. Edges between nodes indicate the relationships of co-purchased or co-commented, reflecting product or user interactions. The labels for these nodes are categorical classes, representing different product categories or post types. We adopt a split setting of 60\% for training, 20\% for validation, and 20\% for testing. More details on the implementations can be found in Appendix~\ref{appendix:implementation}



\subsection{GNN-based Methods (RQ1)}
We conduct experiments using both unimodal and multimodal inputs across various GNN-based models and further evaluate different encoders with multimodal inputs.
Table~\ref{table:encoder_comparison} presents the performance comparison results, and our key findings are as follows.


\begin{table*}[htbp]
\centering
\caption{Node classification accuracy (\%) comparison under different input modality and encoder settings. \textbf{Bold} values denote the highest performance within the same model and dataset setup.}
\label{table:encoder_comparison}
\begin{subtable}[t]{0.5\textwidth}
\centering
\caption{Comparison among various input modality settings. \colorbox[HTML]{C9E3A8}{Green} cells highlight the results of multimodal inputs. }
\label{table:modality}
\resizebox{\textwidth}{!}{%
\begin{tabular}{@{}cccccccc@{}}
\toprule
\toprule
\multicolumn{1}{c}{Model} & Modality & Movies & Toys & Grocery & Arts & CDs & Reddit\\ 
\midrule
\midrule

\multirow{3}{*}{MLP}       
& Text  &  \textbf{46.74} & 74.09 & 79.47 & 82.46 & 44.50 & 59.31	\\
& Image & 38.90 & 49.38 & 58.39 & 68.32 & 47.73 & 73.28 \\
& \cellcolor[HTML]{C9E3A8} Text+Image &  \cellcolor[HTML]{C9E3A8} 45.21 &   \cellcolor[HTML]{C9E3A8} \textbf{74.59} &   \cellcolor[HTML]{C9E3A8} \textbf{84.67} &  \cellcolor[HTML]{C9E3A8} \textbf{84.28} &  \cellcolor[HTML]{C9E3A8} \textbf{52.22} &  \cellcolor[HTML]{C9E3A8} \textbf{76.70} \\

\midrule

\multirow{3}{*}{GCN}       
& Text   & 43.78 &  73.54 &  \textbf{84.44}     &  \textbf{76.93} & 51.45 & 60.49	\\
& Image & 43.32 &  68.23 & 74.77     & 74.48 & 50.93 & 66.06	\\
& \cellcolor[HTML]{C9E3A8} Text+Image  &   \cellcolor[HTML]{C9E3A8} \textbf{46.97} & \cellcolor[HTML]{C9E3A8} \textbf{74.36} & \cellcolor[HTML]{C9E3A8} 79.59     & \cellcolor[HTML]{C9E3A8} 76.76 &  \cellcolor[HTML]{C9E3A8} \textbf{52.68} & \cellcolor[HTML]{C9E3A8} \textbf{65.83} \\

\midrule

\multirow{3}{*}{GraphSAGE} 
& Text  & 43.18 & 75.83 & 85.38 & 85.26  & 52.62  & 68.32\\
& Image  &  \textbf{44.41} & 73.02 & 79.83 & 78.46     & 52.17 & 	76.48 \\
& \cellcolor[HTML]{C9E3A8} Text+Image & \cellcolor[HTML]{C9E3A8} 44.08    &  \cellcolor[HTML]{C9E3A8} \textbf{77.77}  &  \cellcolor[HTML]{C9E3A8} \textbf{86.05} &  \cellcolor[HTML]{C9E3A8} \textbf{85.35}      &  \cellcolor[HTML]{C9E3A8} \textbf{54.75} & \cellcolor[HTML]{C9E3A8} \textbf{78.29} \\

\midrule

\multirow{3}{*}{MMGCN} 
& Text  & 43.84 & 74.87 & 83.84 & 88.82          & 47.71 &  65.73\\
& Image  & 43.20 & 68.82 & 77.26 & 80.18          & 50.45 & 	\textbf{80.99}	\\
& \cellcolor[HTML]{C9E3A8} Text+Image & \cellcolor[HTML]{C9E3A8} \textbf{45.90}    &  \cellcolor[HTML]{C9E3A8} \textbf{75.36}  &  \cellcolor[HTML]{C9E3A8} \textbf{84.63}          & \cellcolor[HTML]{C9E3A8}  \textbf{88.92}  &  \cellcolor[HTML]{C9E3A8} \textbf{51.33} & \cellcolor[HTML]{C9E3A8} 75.58\\

\midrule

\multirow{3}{*}{MGAT} 
& Text  & 39.59 & \textbf{71.10} & 82.95 & \textbf{88.66}          & 52.11 & 67.39 \\
& Image  & 39.52 & 65.55 & 79.18 & 81.03          & 52.85 & \textbf{77.61} \\
& \cellcolor[HTML]{C9E3A8} Text+Image & \cellcolor[HTML]{C9E3A8} \textbf{42.59}    &  \cellcolor[HTML]{C9E3A8} 70.41  &  \cellcolor[HTML]{C9E3A8} \textbf{83.98}          &  \cellcolor[HTML]{C9E3A8} 88.47  &  \cellcolor[HTML]{C9E3A8} \textbf{53.48} & \cellcolor[HTML]{C9E3A8} 75.97\\

\bottomrule
\end{tabular}
}
\end{subtable}
\hfill
\begin{subtable}[t]{0.492\textwidth}
\centering
\caption{Comparison among different enhanced encoders. \colorbox[HTML]{ADC9E9}{Blue} cells indicates the highest within one dataset.}
\label{table:encoder}
\resizebox{\textwidth}{!}{%
\begin{tabular}{@{}cccccccc@{}}
\toprule
\toprule
\multicolumn{1}{c}{Model} & Encoder & Movies & Toys & Grocery & Arts & CDs & Reddit \\ 
\midrule
\midrule

\multirow{3}{*}{MLP}       
& CLIP & 45.21 &  \textbf{74.59} &  \textbf{84.67} &  84.28 &  \textbf{52.22} & 76.70 \\
& CLIP-F & \textbf{47.46} & 73.22 & 82.79 & 82.66 & 48.74  & \textbf{79.77} \\
& CLIP-F-S &  39.73 &  59.88 &  76.94 &  \textbf{84.89} &  49.89  & 79.71 \\

\midrule

\multirow{3}{*}{GCN}       
& CLIP  &   46.97 & \textbf{74.36} & 79.59     & \textbf{76.76} &  \textbf{52.68} & 66.06 \\
& CLIP-F  & \cellcolor[HTML]{ADC9E9} \textbf{47.88} & 74.29 & \textbf{79.67} & 76.52 & 51.47  & 67.67 \\
& CLIP-F-S  &  46.72 &  71.41 &  79.19 &  76.75 &  52.58  & \textbf{67.77} \\

\midrule

\multirow{3}{*}{GraphSAGE} 
& CLIP & 44.08    &  \cellcolor[HTML]{ADC9E9} 
 \textbf{77.77}  &  \cellcolor[HTML]{ADC9E9} 
 \textbf{86.05} &  85.35  &  \cellcolor[HTML]{ADC9E9} \textbf{54.75}  & 76.48 \\
& CLIP-F  & \textbf{46.08} & 76.75 & 85.60 & 84.85 & 53.03  & \textbf{80.03} \\
& CLIP-F-S &  44.92 &  73.44 &  83.84 &  \textbf{86.97} &  54.11  & 79. 94\\

\midrule

\multirow{3}{*}{MMGCN} 
& CLIP & \textbf{45.90}    &  \textbf{75.36}  &  \textbf{84.63}          &  \cellcolor[HTML]{ADC9E9} 
 \textbf{88.92}  &  51.33  & 80.99 \\
& CLIP-F  & 45.42 & 75.22 & 83.61 & 88.22 & 50.61  & 81.71 \\
& CLIP-F-S  &  44.87 &  72.74 &  81.99 &  87.78 &  \textbf{52.18}  & \textbf{81.71} \\

\midrule

\multirow{3}{*}{MGAT} 
& CLIP & 42.59    &  70.41  &  83.98          &  88.47  &  53.48  & 77.61\\
& CLIP-F & \textbf{43.04} & \textbf{71.40} & \textbf{84.46} & 88.41 & 52.01  & 81.33	 \\
& CLIP-F-S &  38.68 &  68.25 &  83.20 &  \textbf{88.68} &  \textbf{55.04}  & \cellcolor[HTML]{ADC9E9} 
 \textbf{81.74} \\

\midrule

\multirow{1}{*}{UniGraph2}       
& CLIP + GNN  &  \textbf{45.91}  &  \textbf{72.70} &  \textbf{80.14} &  \textbf{78.81} &   \textbf{52.13}  &  \textbf{70.64}\\

\bottomrule
\end{tabular}
}
\end{subtable}
\end{table*}

\noindent\textbf{\textit{\ding{72} Finding 1:} Multimodal inputs consistently boost GNN-based model performance over unimodal inputs.}

As the results show in Table~\ref{table:modality}, in most cases, models applying the input of multimodality demonstrate higher performance compared to unimodality inputs. MLP, GraphSAGE, and MMGCN achieve the highest accuracy in 5 out of 6 datasets when using combined text and image inputs. Across all datasets, the multimodal input achieves its highest gains with the MLP model: \textbf{+17.39\%} over text-only on Reddit and \textbf{+26.28\%} over image-only on Grocery. This observation highlights the significant advantages of incorporating multimodal features in MMGL tasks and reinforces their effectiveness in enhancing the performance of GNN-based models.

\noindent\textbf{\textit{\ding{72} Finding 2:} Modality and structure alignment enhancement methods do not consistently enable GNN-based models to effectively handle MMGL tasks.}

From Table~\ref{table:encoder}, we observe that the structure-aware encoder \textit{CLIP-F-S} improves node classification accuracy for MLP, GraphSAGE, MMGACN, and MGAT models on the Arts and CD datasets. However, the highest accuracy across all datasets occurs in GCN, GraphSAGE, and MMGCN when using the CLIP and CLIP-F encoders, both of which are non-structure-aware. Notably, Notably, in the Reddit dataset, which features high-quality visual images but short textual captions, all modality-enhanced encoders lead to performance gains. This highlights the need for advanced alignment techniques that can more effectively integrate multiple node attribute modalities with graph structural information.



\subsection{LLM-based Methods (RQ2)}
We compare the performance of original text-only inputs with the summary of image and text inputs, where the MLLM is applied as a modality aligner. To assess the impact of structural awareness, we include a baseline that concatenates node image and text embeddings as input for GraphLLMs that accept embedding features. Results are presented in Table~\ref{table:aligner} and Figure~\ref{fig:augmentor}. Here are three key findings.


\begin{table}[htbp]
    \begin{minipage}{0.65\textwidth}
        \centering
        \resizebox{\textwidth}{!}{
        \begin{tabular}{@{}cccccccc@{}}
        \toprule
        \toprule
        \multicolumn{1}{c}{Model} & Input & Movies & Toys & Grocery & Arts & CDs & Reddit \\ 
        \midrule
        \midrule
        
        \multirow{2}{*}{GraphPrompter}       
        & Text & \textbf{46.36} & 74.08 & 85.00 & \textbf{84.41} & 46.27 & 76.95 \\
        & Text and Image & 45.55 & \textbf{76.56} & \textbf{85.47} & 64.87 & \textbf{52.06} & - \\
        
        \midrule
        
        \multirow{2}{*}{LLaGA}       
        & Text & \textbf{49.57} & \textbf{77.51} & \cellcolor[HTML]{ADC9E9} \textbf{86.25} & \textbf{89.32} & 53.12 & 74.41\\
        & Text and Image & 48.73 & 74.97 & 86.16 & 76.34 & \cellcolor[HTML]{ADC9E9} \textbf{54.45} & \cellcolor[HTML]{ADC9E9} \textbf{82.85} \\
        
        \midrule
        
        \multirow{2}{*}{GraphGPT}       
        & Text & \textbf{10.19} & \textbf{37.38} & 59.76 & 57.35 & 23.86 & 21.83\\
        & Text and Image & 8.82 & 36.10 & \textbf{59.87} & \textbf{58.73} & \textbf{29.57} & \textbf{26.03} \\
        
        \midrule
        
        \multirow{2}{*}{GraphTranslator}       
        & Text & \textbf{7.11} & 11.96 & \textbf{16.28} & \textbf{33.00} & 9.35 & - \\
        & Text and Image & 5.34 & \textbf{19.96} & 13.68 & - & - & - \\
        
        \midrule
        
        \multirow{1}{*}{MLaGA}       
        & Text and Image & \cellcolor[HTML]{ADC9E9} \textbf{49.78} & \cellcolor[HTML]{ADC9E9} \textbf{80.00} &  \textbf{85.78} & \cellcolor[HTML]{ADC9E9} \textbf{89.79} & \cellcolor[HTML]{ADC9E9} \textbf{54.45} & - \\
        
        \bottomrule
        \end{tabular}
        }
        \caption{Node classification accuracy (\%) comparison between original text and image-to-text augmented inputs. \textbf{Bold} denotes the highest within the same model and dataset setup. \colorbox[HTML]{ADC9E9}{Blue} cells indicate the highest within one dataset. -- means omitted results due to time efficiency constraints.}
        \label{table:aligner}
    \end{minipage}
    \hfill
    \begin{minipage}{0.33\textwidth}
        \centering
        \includegraphics[width=\linewidth]{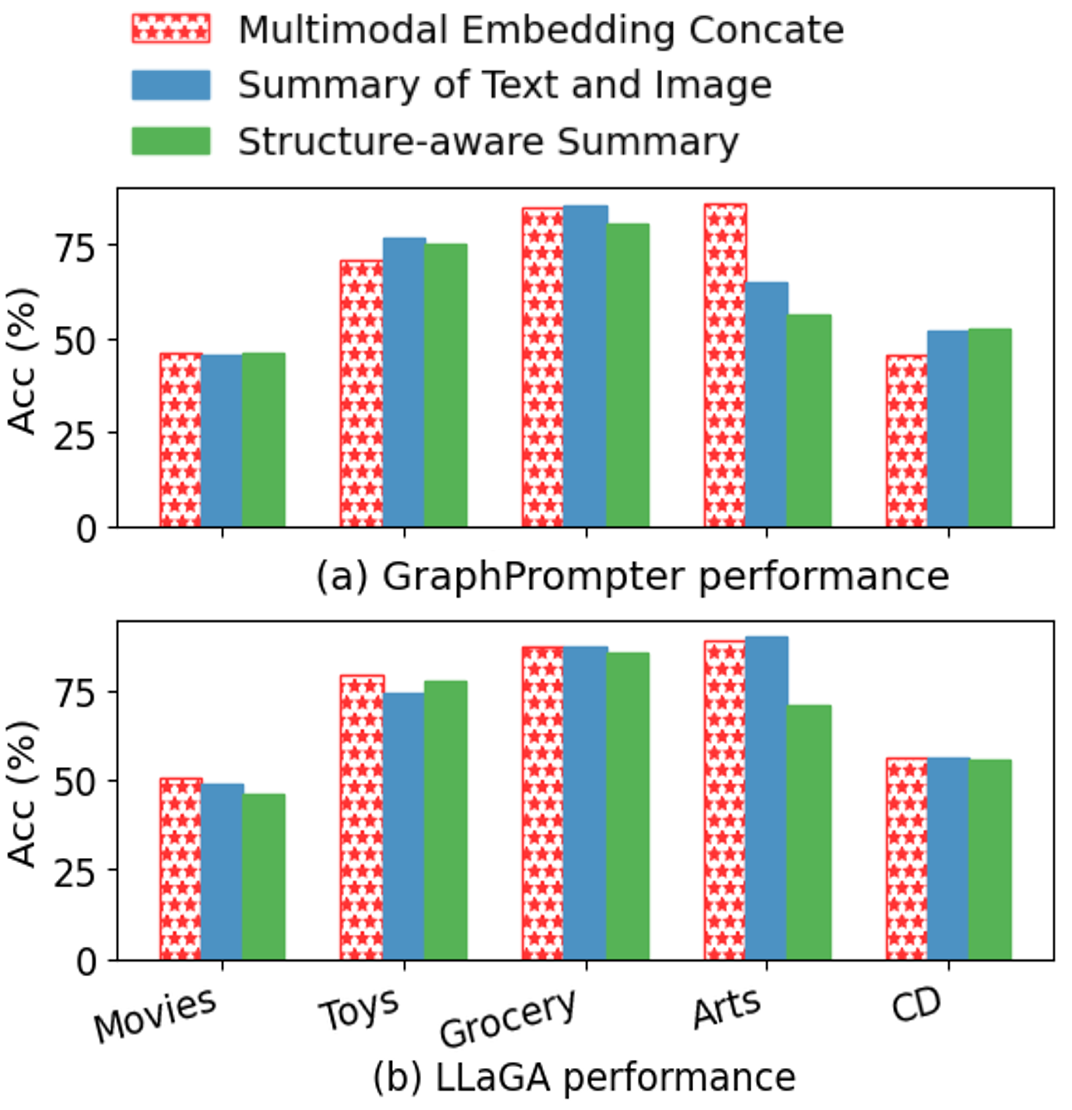}
        \captionof{figure}{Different aligner augmentation methods comparison}
        \label{fig:augmentor}
    \end{minipage}
\end{table}

\noindent\textbf{\textit{\ding{72} Finding 3:} The performance of aligners exhibits model-specific sensitivity.}

An analysis of results across different models reveals a clear contrast in the effectiveness of image-to-text augmentation. While GraphGPT and GraphPrompter show consistent performance gains in 4 out of 6 and 3 out of 5 datasets, respectively, with aligner augmentation, LLaGA exhibits performance degradation in 4 out of 6 datasets when the aligner is applied. 
This discrepancy underscores the strong dependency of the aligner augmentation effectiveness on the underlying GraphLLM architecture. GraphPrompter and GraphGPT heavily rely on prompt information, benefiting from modality transformation as visual information is effectively summarized within the prompt, making it easier for the model to interpret. Conversely, LLaGA places more emphasis on embedding quality, where the alignment of textual and visual modalities through a well-pretrained modality-alignment encoder becomes crucial. As shown in Figure~\ref{fig:augmentor}, when the aligner augments multimodality in the embedding space, LLaGA shows substantial performance improvements.

\noindent\textbf{\textit{\ding{72} Finding 4:} Image-to-Text alignment methods demonstrate dataset-specific sensitivity.}

From a dataset perspective, augmentation methods show varying sensitivity to dataset characteristics. Specifically, aligned multimodal inputs yield performance gains of 5.79\%, 5.71\%, and 1.33\% on the CDs dataset when using GraphPrompter, GraphGPT and LLaGA, respectively. While most augmentation methods improve performance on the CDs dataset across different models, none demonstrate improvement on the Movies dataset. This dataset-specific behavior indicates that the effectiveness of augmentation strategies is closely tied to intrinsic dataset properties such as data complexity, modality alignment, and feature distribution. Taken together with Finding 3, these results highlight the importance of designing tailored augmentation strategies that account for both model architecture and dataset characteristics to fully leverage multimodal information.

\noindent\textbf{\textit{\ding{72} Finding 5:} MLaGA achieves state-of-the-art performance, while structure-aware augmentation during data pre-processing provides limited or inconsistent benefits.}

MLaGA delivers the highest accuracy on 4 out of 5 datasets, owing to its integrated design that jointly models structural and multimodal information. However, incorporating structural information during data pre-processing yields less promising results.

As shown in Figure~\ref{fig:augmentor}, a comparison between structure-aware and non-structure-aware augmentation reveals that prompt-level structural augmentation does not consistently improve performance and can even degrade it due to the introduction of noise. In GraphPrompter, only marginal improvements are observed on the Movies (0.57\%) and CDs (0.52\%) datasets. In contrast, LLaGA experiences performance drops across all datasets when structure-aware augmentation is applied. These findings suggest that incorporating structural awareness at the prompt level during pre-processing does not guarantee performance gains and may, in some cases, be detrimental.


\begin{table*}[htbp] 
\centering

\small
\caption{Node classification accuracy (\%) comparison of various MLLMs on all datasets. \textbf{Bold} values denote the highest performance among those utilized within the same MLLM backbone model and dataset setup. \colorbox[HTML]{C9E3A8}{green} cells indicate fine-tuning and \colorbox[HTML]{F7C7AD}{pink} cells denote in-context learning.}

\label{table:predictor} 

\resizebox{\textwidth}{!}{

\begin{tabular}{@{}lcccccccl@{}}
\toprule
\toprule
Model & Fine-tuned & Structure Data & Movies & Toys & Grocery & Arts & CDs & Reddit \\ 
\midrule
\midrule

\multirow{3}{*}{LLaVA-1.5-7B} 

& No & None & 13.19 & 47.52 & 57.88 & 72.96 & 41.30 & 34.82\\
& \cellcolor[HTML]{C9E3A8} Yes & \cellcolor[HTML]{C9E3A8} None  & \cellcolor[HTML]{C9E3A8} \textbf{45.58} & \cellcolor[HTML]{C9E3A8} 74.32 & \cellcolor[HTML]{C9E3A8} 85.89 & \cellcolor[HTML]{C9E3A8} 84.47 & \cellcolor[HTML]{C9E3A8} 53.55 & \cellcolor[HTML]{C9E3A8} 82.45\\
& \cellcolor[HTML]{C9E3A8} Yes & \cellcolor[HTML]{C9E3A8} Neighbor Text & \cellcolor[HTML]{C9E3A8} 44.65 & \cellcolor[HTML]{C9E3A8} \textbf{74.82} & \cellcolor[HTML]{C9E3A8} \textbf{85.91} & \cellcolor[HTML]{C9E3A8} \textbf{82.94} & \cellcolor[HTML]{C9E3A8} \textbf{53.73} & \cellcolor[HTML]{C9E3A8} \textbf{82.76} \\

\midrule

\multirow{8}{*}{Qwen-VL-7B} 

& No & None & 12.56 & 48.90 & 43.73 & 67.69 & 37.51 & 47.52 \\
& \cellcolor[HTML]{F7C7AD} No & \cellcolor[HTML]{F7C7AD} Neighbor Text+Image & \cellcolor[HTML]{F7C7AD} 29.18 & \cellcolor[HTML]{F7C7AD} 57.04 & \cellcolor[HTML]{F7C7AD} 66.21 & \cellcolor[HTML]{F7C7AD} 68.66  & \cellcolor[HTML]{F7C7AD} 41.64 & \cellcolor[HTML]{F7C7AD} 56.05 \\
& \cellcolor[HTML]{C9E3A8} Yes & \cellcolor[HTML]{C9E3A8} None & \cellcolor[HTML]{C9E3A8} \textbf{54.18} & \cellcolor[HTML]{C9E3A8} 80.38  & \cellcolor[HTML]{C9E3A8} 88.33  & \cellcolor[HTML]{C9E3A8} \textbf{92.43 } & \cellcolor[HTML]{C9E3A8} \textbf{58.99}  & \cellcolor[HTML]{C9E3A8} 85.76 \\
& \cellcolor[HTML]{C9E3A8} Yes & \cellcolor[HTML]{C9E3A8} Neighbor Text & \cellcolor[HTML]{C9E3A8} 51.66  & \cellcolor[HTML]{C9E3A8} \textbf{80.96 } & \cellcolor[HTML]{C9E3A8} \textbf{88.34 } & \cellcolor[HTML]{C9E3A8} 92.04  & \cellcolor[HTML]{C9E3A8} 58.81  & \cellcolor[HTML]{C9E3A8} \textbf{85.96} \\
& \cellcolor[HTML]{C9E3A8} Yes & \cellcolor[HTML]{C9E3A8} Neighbor Image & \cellcolor[HTML]{C9E3A8} 46.27  & \cellcolor[HTML]{C9E3A8} 80.12  & \cellcolor[HTML]{C9E3A8} 87.78 & \cellcolor[HTML]{C9E3A8} 91.13 & \cellcolor[HTML]{C9E3A8} 57.45  & \cellcolor[HTML]{C9E3A8} 84.97\\
& \cellcolor[HTML]{C9E3A8} Yes & \cellcolor[HTML]{C9E3A8} Neighbor Text+Image & \cellcolor[HTML]{C9E3A8} 45.88  & \cellcolor[HTML]{C9E3A8} 79.80  & \cellcolor[HTML]{C9E3A8} 87.78 & \cellcolor[HTML]{C9E3A8} 91.15  & \cellcolor[HTML]{C9E3A8} 57.01  & \cellcolor[HTML]{C9E3A8} 85.25 \\ 




\bottomrule
\end{tabular}
}
\end{table*}


\subsection{MLLM-based Methods (RQ3)}

Table~\ref{table:predictor} presents the results for in-context learning and finetuned prediction methods. LLaVA-1.5-7B supports only a single image as input, so we conduct more structure-aware experiments using the Qwen-VL-7B model. The following are two key findings for this section:

\noindent\textbf{\textit{\ding{72} Finding 6:} MLLM prediction performance is initially low on MMGL, but finetuning significantly boosts accuracy.} 

The results show low accuracy for both LLaVA and QWen-VL under in-context learning settings, but fine-tuning leads to significant performance improvements across all datasets. Specifically, with Qwen-VL-7B in the Movies dataset, the initial precision with no fine-tuning and no structure data is only 12.56\%, but fine-tuning boosts accuracy to 54.18\%, achieving an impressive increase by \textbf{4.3$\times$}. LLaVA-1.5-7B exhibits a similarly rapid increase, with a \textbf{3.5$\times$} improvement. As highlighted by the bold values indicating the highest accuracy for each MLLM model, fine-tuned predictors yield substantial performance improvements. In particular, Qwen-VL-7B achieves gains of \textbf{+41.62\%}, \textbf{+32.06\%}, \textbf{+44.61\%}, \textbf{+24.74\%}, \textbf{+21.64\%}, and \textbf{+38.44\%} on the Movies, Toys, Grocery, Arts, CDs, and Reddit datasets, respectively, compared to the initial baseline. These improvements reflect MLLMs' strong multimodal alignment and understanding capabilities, highlighting their potential as backbones for MMGL tasks and the need for further exploration.

\noindent\textbf{\textit{\ding{72} Finding 7:} While structure-aware strategies are necessary, injecting them into MLLM prompts is not sufficient.}

Structure-aware fine-tuning methods demonstrate consistent and stable improvements over non-structure fine-tuning across all datasets in LLaVA-1.5-7B and 3 out of 6 datasets in Qwen-VL-7B. This highlights the potential benefits of incorporating structural information during finetuning. However, the performance gains are modest, with less than 1\% improvement compared to non-structure fint-tuning results. Moreover, in some cases, adding structure awareness can even decrease accuracy, particularly when leveraging the visual information of neighbors. These results emphasize the need for more advanced methods to seamlessly integrate graph structure information into MLLMs.


\subsection{Mixture-of-All (RQ4)}
After evaluating each MMGL paradigm and the respective models, we combine all the paradigms together to comprehensively assess and summarize the potential of MLLMs for MMGL.

\begin{figure}[htb]
\centering
\begin{minipage}{0.57\textwidth}
    \includegraphics[width=\linewidth]{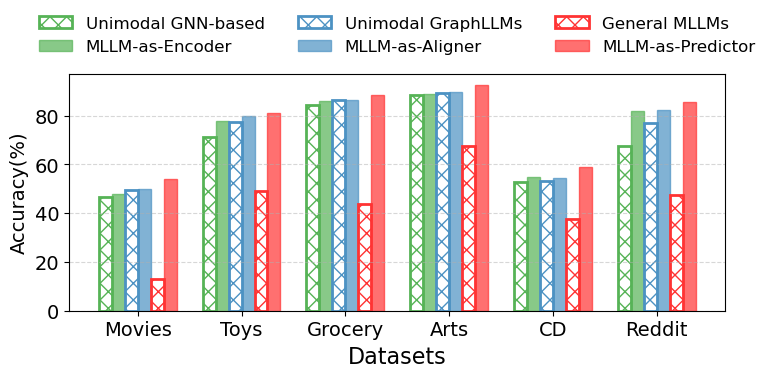}
    \caption{Performance w/o MLLM enhancement.}
    \label{fig:bestperformance}
\end{minipage}
\hfill
\begin{minipage}{0.42\textwidth}
    \includegraphics[width=\linewidth]{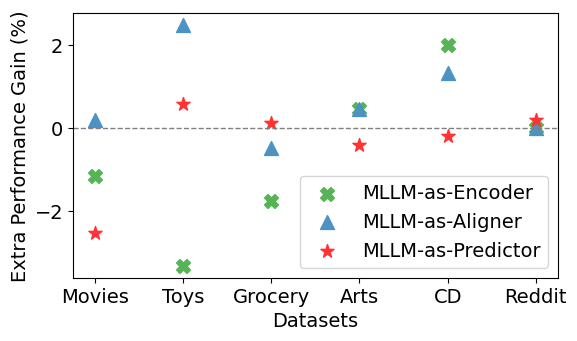}
    \caption{Extra performance gain of structure-aware to non-structure methods.}
    \label{fig:structuregain}
\end{minipage}

\end{figure}

We compare the best performance of the original methods with MLLM-enhanced models, highlighting the performance gains achieved through MLLM integration. The results are illustrated in Figure~\ref{fig:bestperformance}. 
To evaluate the effectiveness of structure awareness across all approaches, we assess the extra performance gain provided by structure-aware methods. This performance gain is calculated as the difference between the highest accuracy achieved by structure-aware methods and the highest by non-structure methods. The results are presented in Figure~\ref{fig:structuregain}.

\noindent\textbf{\textit{\ding{72} Finding 8:} MLLM-as-Predictor demonstrates the greatest potential for MMGL, while both MLLM-as-Encoder and MLLM-as-Aligner generally lead to performance improvements.}

Without the involvement of MLLMs, GraphLLMs consistently outperform both GNN-based methods and general-purpose MLLMs across all datasets, highlighting the advantages of integrating language models with graph structures. Although general-purpose MLLMs consistently perform the worst among all methods, they achieve top performance across all datasets after fine-tuning. This substantial performance gap underscores the strong potential of the MLLM-as-Predictor strategy for MMGL. Meanwhile, both the MLLM-as-Encoder and MLLM-as-Aligner approaches consistently lead to performance improvements. Notably, while the state-of-the-art GNN-based method UniGraph2 and the LLM-based method MLaGA demonstrate high accuracy, fine-tuned MLLMs without structural data can match or even surpass their performance on certain datasets. This further highlights the potential for an MLLM-based method for MMGL.

\noindent\textbf{\textit{\ding{72} Finding 9:} The effectiveness of structure awareness for different MMGL methods relies heavily on data properties in MMGs.}

Both graph structure density and modality quality significantly affect the effectiveness of structure awareness. Results reveal that for datasets with higher average degrees, particularly in the CDs dataset with a high average degree of 47, structure awareness consistently improves performance across all three approaches. In contrast, in sparse graphs such as Toys, structure awareness not only fails to provide benefits but may even degrade performance. Nevertheless, the modality aligner helps enrich node representations, thereby improving predictive performance. Additionally, the Movies dataset, with the lowest image quality (see Table~\ref{table:modality}), demonstrates how poor image features can undermine the structural effectiveness of all methods. These findings highlight the need to consider sparse graph structures and imbalanced multimodal quality when dealing with MMGL.

\section{Conclusion}
\label{sec:conclusion}
In this paper, we present Graph-MLLM, a comprehensive benchmark for multimodal graph learning (MMGL) that evaluates state-of-the-art methods across six diverse datasets. Existing MMGL approaches are grouped into three paradigms based on how multimodal large language models (MLLMs) are utilized: MLLM-as-Encoder for GNN-based methods, MLLM-as-Aligner for LLM-based methods, and MLLM-as-Predictor, where MLLMs are used as standalone predictors. Extensive experiments show that fine-tuning MLLMs as predictors yields the most significant performance gains, even without explicit graph structural information, highlighting their potential as powerful backbones for MMGL. We hope that Graph-MLLM serves as a fair and standardized benchmark to facilitate future research and foster continued innovation in this rapidly evolving field.

\bibliographystyle{plainnat} 
\bibliography{main}










\newpage
\appendix

\newpage
\section{Appendix}
\label{sec: appendix}

\subsection{Dataset Statistics}
\label{appendix:dataset}
Our experimental benchmark comprises six datasets spanning two distinct domains, with detailed statistical characteristics presented in Table~\ref{table:dataset}.

\begin{table}[htbp]
    \centering
    \caption{Statistics of the datasets.}
    \small
    \begin{tabular}{lccccc}
        \hline
        Dataset & \#Nodes & \#Edges & Avg.\#Degree &\#Classes &Domain \\
        \hline
        Movies  &16,672&218,390& 26 &19 & E-commerce\\
        Toys &20,695&126,886& 12 &18 & E-commerce \\
        Grocery &  84,379      &   693,154      &     16         &    20   & E-commerce  \\
        Arts  & 28,195 & 197,428 & 14 & 7 & E-commerce\\
        CDs &  36,272       &   844,878      &       47       &     15  & E-commerce  \\
        Reddit &  99,638     &    1,167,188      &    23     &     50  & Social Network  \\
        \hline
    \end{tabular}
    \label{table:dataset}
\end{table}

\subsection{Implementations}
\label{appendix:implementation}

A diverse set of state-of-the-art (SOTA) graph learning methods is involved in this benchmark: 
\begin{itemize}
\item
For GNN-based methods, we evaluate the conventional GCN~\cite{kipf2016gcn}, the widely used GraphSAGE~\cite{hamilton2017graphSAGE}, and MLP~\cite{rosenblatt1958perceptron} as a non-graph baseline. We also include MMGCN~\cite{mmgcn} and MGAT~\cite{tao2020mgat}, which are specifically designed for multimodal graphs. To ensure fair comparisons, all models are evaluated under consistent experimental settings, and our implementation follows the MM-Bench framework~\cite{zhu2024multimodalgraphbenchmark}. Moreover, we adopt the state-of-the-art UniGraph2~\cite{uniGraph2}. However, due to the $O(n^3)$ time complexity of the shortest path distance (SPD) computation in the original implementation, we exclude the SPD module for efficiency in our evaluation.
\item
In the category of LLM-based methods, we evaluate several GraphLLMs with diverse architectures, including GraphPrompter~\cite{liu2024graphPrompter}, LLaGA~\cite{chen2024llaga}, GraphGPT~\cite{tang2024graphgpt}, GraphTranslator~\cite{zhang2024graphtranslator}, as well as the state-of-the-art MLaGA~\cite{mlaga2025}. All evaluations strictly follow the official implementations.
\item 
Beyond these, we include the SOTA multimodal models QWen-VL~\cite{bai2023qwen} and LLaVA-1.5-7B~\cite{liu2023llava} for predictor experiments, with QWen-VL also serving as the aligner for RQ2. We adhere to the official guidelines for zero-shot evaluation and fine-tuning strictly.  For all models requiring embedding inputs, we utilize the pre-trained CLIP~\cite{clip} model as the encoder to effectively process multimodal features. 
\end{itemize}
More detailed implementation and resource settings can be found in the Graph-MLLM repository: \url{https://github.com/oamyjin/Graph-MLLM}.

\subsection{Aligner Prompt design}
\label{appendix:prompt}
\subsubsection{Image summary} We use Qwen-VL\cite{bai2023qwen} to generate the image summaries as shown in Table~\ref{table:augmentor_image_prompts}

\begin{table*}[htbp]
    \centering
    \caption{describes the prompts used to generate a text description of the image by MLLM}
    \label{table:augmentor_image_prompts}
    \resizebox{1\textwidth}{!}{%
        \begin{tabular}{p{1.5cm} p{12cm}}
            \toprule
            \textbf{Movies:} & \textcolor{red}{<image input>} Given an image of a \textcolor{blue}{movie} from the \textcolor{blue}{Amazon movies dataset }, generate a concise and detailed summary. Focus on describing key visual concepts. Ensure the summary is informative and useful for understanding the product as described in user reviews, without losing critical details or introducing unnecessary information. \\
            \midrule
            \textbf{Toys:} & \textcolor{red}{<image input>} Given an image of a \textcolor{blue}{toy} from the \textcolor{blue}{Amazon toys dataset }, generate a concise and detailed summary. Focus on describing key visual concepts. Ensure the summary is informative and useful for understanding the product as described in user reviews, without losing critical details or introducing unnecessary information. \\
            \midrule
            \textbf{Grocery:} & \textcolor{red}{<image input>} Given an image of a \textcolor{blue}{grocery} from the \textcolor{blue}{Amazon grocery dataset }, generate a concise and detailed summary. Focus on describing key visual concepts. Ensure the summary is informative and useful for understanding the product as described in user reviews, without losing critical details or introducing unnecessary information. \\
            \midrule
            \textbf{CDs:} & \textcolor{red}{<image input>} Given an image of a \textcolor{blue}{CD} from the \textcolor{blue}{Amazon CD dataset }, generate a concise and detailed summary. Focus on describing key visual concepts. Ensure the summary is informative and useful for understanding the product as described in user reviews, without losing critical details or introducing unnecessary information. \\
            \midrule
            \textbf{Arts:} & \textcolor{red}{<image input>} Given an image of an \textcolor{blue}{artwork} from the \textcolor{blue}{Amazon Art dataset }, generate a concise and detailed summary. Focus on describing key visual concepts. Ensure the summary is informative and useful for understanding the product as described in user reviews, without losing critical details or introducing unnecessary information \\
            \midrule
            \textbf{Reddit:} & \textcolor{red}{<image input>} Given an image of a \textcolor{blue}{post} from the \textcolor{blue}{Reddit dataset}, generate a concise and detailed summary. Focus on describing key visual concepts. Ensure the summary is informative and useful for understanding the post as described in the caption, without losing critical details or introducing unnecessary information. \\
            \bottomrule
        \end{tabular}
    }
\end{table*}

\subsection{Non-structure-aware Modality Synthesis}  
We synthesize the original textual information with generated image summaries using an LLM, following the prompts below. The prompts are shown in Table~\ref{table :non_structure_aware_modality_synthesis}.

\begin{table*}[htbp]
    \centering
    \caption{shows the synthesis prompts for the non-structure-aware aligner case.}
    \label{table :non_structure_aware_modality_synthesis}
    \resizebox{1\textwidth}{!}{%
        \begin{tabular}{p{1.5cm} p{12cm}}
            \toprule
            \textbf{Movies:} &Given the text information of a product from the \textcolor{blue}{Amazon Movies} dataset: \textcolor{red}{<text information>}. Image summary: \textcolor{red}{<image summary>} Questions: Using the title, description, and image summary of the product provided above, create an informative and concise description that effectively highlights the product's key features. \\
            \midrule
            \textbf{Toys:} & Given the text information of a product from the \textcolor{blue}{Amazon toys} dataset: \textcolor{red}{<text information>}. Image summary: \textcolor{red}{<image summary>} Questions: Using the title, description, and image summary of the product provided above, create an informative and concise description that effectively highlights the product's key features. \\ 
            \midrule
            \textbf{Grocery:} & Given the text information of a product from the \textcolor{blue}{Amazon grocery} dataset: \textcolor{red}{<text information>}. Image summary: \textcolor{red}{<image summary>} Questions: Using the title, description, and image summary of the product provided above, create an informative and concise description that effectively highlights the product's key features. \\ 
            \midrule
            \textbf{CDs:} & Given the text information of a product from the \textcolor{blue}{Amazon CD} dataset: \textcolor{red}{<text information>}. Image summary: \textcolor{red}{<image summary>} Questions: Using the title, description, and image summary of the product provided above, create an informative and concise description that effectively highlights the product's key features.\\
            \midrule
            \textbf{Arts:} & Given the text information of a product from the \textcolor{blue}{Amazon Art} dataset: \textcolor{red}{<text information>}. Image summary: \textcolor{red}{<image summary>} Questions: Using the title, description, and image summary of the product provided above, create an informative and concise description that effectively highlights the product's key features.\\
            \midrule
            \textbf{Reddit:} & Given the text information of a post from the \textcolor{blue}{Reddit} dataset: \textcolor{red}{<text information>}. Image summary: \textcolor{red}{<image summary>} Questions: Using the caption and image summary of the post provided above, create an informative and concise description that effectively highlights the post's key features.\\
            \bottomrule
        \end{tabular}
    }
\end{table*}

\subsection{Structure-Aware Modality Synthesis}
To incorporate structure-aware modality information, we design our prompts for modality synthesis as shown in Table~\ref{table: modality_synthesis_structure_aware}.

\begin{table*}[htbp]
    \centering
    \caption{shows the synthesis prompts for the structure-aware aligner case.}
    \label{table: modality_synthesis_structure_aware}
       \resizebox{1\textwidth}{!}{%
            \begin{tabular}{p{1.5cm} p{12cm}}
            \toprule
            \textbf{Movies:} &Given the text information of a product from the \textcolor{blue}{Amazon movies} dataset: \textcolor{red}{<text information>}. Image summary: \textcolor{red}{<image summary>}. Also given the information of co-purchased or co-reviewed products: text information: \textcolor{red}{<neighbor text information>}, image summary: \textcolor{red}{<neighbor image summary>} (or, if unavailable: 'No co-purchased or co-reviewed product information is available.') Questions: Using the product's title, description, and image summary provided above, along with any co-purchase or co-review data, generate a concise yet informative description of the product.  \\
            \midrule
            \textbf{Toys:} & Given the text information of a product from the \textcolor{blue}{Amazon toys} dataset: \textcolor{red}{<text information>}. Image summary: \textcolor{red}{<image summary>}. Also given the information of co-purchased or co-reviewed products: text information: \textcolor{red}{<neighbor text information>}, image summary: \textcolor{red}{<neighbor image summary>} (or, if unavailable: 'No co-purchased or co-reviewed product information is available.') Questions: Using the product's title, description, and image summary provided above, along with any co-purchase or co-review data, generate a concise yet informative description of the product. \\
            \midrule
            \textbf{Grocery:} & Given the text information of a product from the \textcolor{blue}{Amazon grocery} dataset: \textcolor{red}{<text information>}. Image summary: \textcolor{red}{<image summary>}. Also given the information of co-purchased or co-reviewed products: text information: \textcolor{red}{<neighbor text information>}, image summary: \textcolor{red}{<neighbor image summary>} (or, if unavailable: 'No co-purchased or co-reviewed product information is available.') Questions: Using the product's title, description, and image summary provided above, along with any co-purchase or co-review data, generate a concise yet informative description of the product. \\
            \midrule
            \textbf{CDs:} & Given the text information of a product from the \textcolor{blue}{Amazon CD} dataset: \textcolor{red}{<text information>}. Image summary: \textcolor{red}{<image summary>}. Also given the information of co-purchased or co-reviewed products: text information: \textcolor{red}{<neighbor text information>}, image summary: \textcolor{red}{<neighbor image summary>} (or, if unavailable: 'No co-purchased or co-reviewed product information is available.') Questions: Using the product's title, description, and image summary provided above, along with any co-purchase or co-review data, generate a concise yet informative description of the product.\\
            \midrule
            \textbf{Arts:} & Given the text information of a product from the \textcolor{blue}{Amazon Art} dataset: \textcolor{red}{<text information>}. Image summary: \textcolor{red}{<image summary>}. Also given the information of co-purchased or co-reviewed products: text information: \textcolor{red}{<neighbor text information>}, image summary: \textcolor{red}{<neighbor image summary>} (or, if unavailable: 'No co-purchased or co-reviewed product information is available.') Questions: Using the product's title, description, and image summary provided above, along with any co-purchase or co-review data, generate a concise yet informative description of the product.\\
            \midrule
            \textbf{Reddit:} & Given the text information of a post from the \textcolor{blue}{Reddit} dataset: \textcolor{red}{<text information>}. Image summary: \textcolor{red}{<image summary>}. Also given the information of co-commented posts: text information: \textcolor{red}{<neighbor text information>}, image summary: \textcolor{red}{<neighbor image summary>} (or, if unavailable: 'No co-commented post information is available.') Questions: Using the post's caption and image summary provided above, along with any co-commented data, generate a concise yet informative description of the post.\\
            \bottomrule
        \end{tabular}
    }
\end{table*}

\subsection{MLLM as Predictor Prompt Design}
To directly enable MLLM as the predictor, we design the prompt accordingly as shown in Table~\ref{table: MLLM_as_predictor_prompt} and Table~\ref{table: MLLM_as_predictor_prompt2}. In fine-tuning strategies, the prompt includes \textit{'Assistant: <true label>'}, while in context-learning cases, it is excluded.

\begin{table*}[htbp]
    \centering
    \caption{shows the synthesis prompts for the non-structure-aware predictor case.}
    \label{table: MLLM_as_predictor_prompt}
    \resizebox{1\textwidth}{!}{%
        \begin{tabular}{p{2.5cm} p{12.5cm}}
            \toprule
            \textbf{Movies / Toys / Grocery / CDs / Arts} &Given the target product information on Amazon:
            Picture: \textcolor{red}{<image input>}
            Title and description: \textcolor{red}{<text information>}. 
            Question: Based on the target product's picture, title, and description, which category does the target product belong to? Choose from the following options: \textcolor{red}{<candidates set>}.
            
            Assistant: \textcolor{red}{<truth label>}
            \\
            
            \midrule
                    \textbf{Reddit} &Given the target post information on Reddit:
            Picture: \textcolor{red}{<image input>}
            Caption: \textcolor{red}{<text information>}.
            Question: Based on the target post's picture and caption, which category does the target post belong to? Choose from the following options: \textcolor{red}{<candidates set>}.
            
            Assistant: \textcolor{red}{<truth label>} \\
            \bottomrule
        \end{tabular}
    }
\end{table*}

\begin{table*}[htbp]
    \centering
    \caption{shows the synthesis prompts for the structure-aware predictor case.}
    \label{table: MLLM_as_predictor_prompt2}
    \resizebox{1\textwidth}{!}{%
        \begin{tabular}{p{2.5cm} p{12.5cm}}
            \toprule
            \textbf{Movies / Toys / Grocery / CDs / Arts} &Given the target product information on Amazon:
            Picture: \textcolor{red}{<image input>}
            Title and description: \textcolor{red}{<text information>}.
            Co-purchased or co-reviewed products: Picture1: \textcolor{red}{<image input>}; Title1: \textcolor{red}{<text information>} ; Picture2: \textcolor{red}{<image input>}; Title2: \textcolor{red}{<text information>} ; Picture3: \textcolor{red}{<image input>}; Title3: \textcolor{red}{<text information>}. 
            Question: Based on the target product's picture, title, description, and related products, which category does the target product belong to? Choose from the following options: \textcolor{red}{<candidates set>}.
            
            Assistant: \textcolor{red}{<truth label>}
            \\
            
            \midrule
                    \textbf{Reddit} &Given the target post information on Reddit:
            Picture: \textcolor{red}{<image input>}
            Caption: \textcolor{red}{<text information>}.
            Co-commented posts: Picture1: \textcolor{red}{<image input>}; Caption1: \textcolor{red}{<text information>} ; Picture2: \textcolor{red}{<image input>}; Caption2: \textcolor{red}{<text information>} ; Picture3: \textcolor{red}{<image input>}; Caption3: \textcolor{red}{<text information>}.
            Question: Based on the target post's picture, caption, and related posts, which category does the target post belong to? Choose from the following options: \textcolor{red}{<candidates set>}.
            
            Assistant: \textcolor{red}{<truth label>} \\
            \bottomrule
        \end{tabular}
    }
\end{table*}

\end{document}